\documentclass[conference]{IEEEtran}
\IEEEoverridecommandlockouts
\usepackage{cite}
\usepackage{amsmath,amssymb,amsfonts}
\usepackage{algorithmic}
\usepackage{graphicx}
\usepackage{textcomp}
\usepackage{xcolor}
\usepackage{epsfig}
\usepackage{algorithm}
\usepackage{algorithmic}

\DeclareMathOperator{\sign}{sign}

\begin{document}

\title{AuxBlocks: Defense Adversarial Example via Auxiliary Blocks\\
\thanks{The work is partially supported by Project No. xxxxxx.}
}

\author{\IEEEauthorblockN{Yueyao Yu, Pengfei Yu and Wenye Li}
\IEEEauthorblockA{\textit{The Chinese University of HongKong,Shenzhen} \\
yueyaoyu@link.cuhk.edu.cn, pengfeiyu@link.cuhk.edu.cn, wyli@cuhk.edu.cn}
}

\maketitle

\begin{abstract}
Deep learning models are vulnerable to adversarial examples, which poses an indisputable threat to their applications. However, recent studies observe gradient-masking defenses are self-deceiving methods if an attacker can realize this defense. In this paper, we propose a new defense method based on appending information. We introduce the Aux Block model to produce extra outputs as a self-ensemble algorithm and analytically investigate the robustness mechanism of Aux Block. We have empirically studied the efficiency of our method against adversarial examples in two types of white-box attacks, and found that even in the full white-box attack where an adversary can craft malicious examples from defense models, our method has a more robust performance of about 54.6\% precision on Cifar10 dataset and 38.7\% precision on Mini-Imagenet dataset. Another advantage of our method is that it is able to maintain the prediction accuracy of the classification model on clean images, and thereby exhibits its high potential in practical applications.
\end{abstract}


\section{Introduction}

Breakthroughs in deep neural networks (DNN) learning have achieved exceptional performance across a wide range of applications, such as in image classification \cite{krizhevsky2012imagenet} and in speech recognition \cite{hinton2012deep}. However, recent studies showed that neural networks are vulnerable to adversarial examples, and attackers can design maliciously perturbed inputs to mislead a model at the test phase \cite{biggio2013evasion, szegedy2013intriguing, kurakin2016adversarial}. Furthermore, these inputs can transfer across different models, and one example can attack several models without knowing model structures or parameters \cite{szegedy2013intriguing}. This threats DNN applications in security-sensitive systems such as in self-driving cars and in identity authentication.

\begin{figure*}
\centering
\includegraphics[width=1.5cm,height=1.5cm]{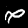}
\includegraphics[width=1.5cm,height=1.5cm]{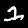}
\includegraphics[width=1.5cm,height=1.5cm]{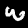}
\includegraphics[width=1.5cm,height=1.5cm]{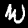}
\includegraphics[width=1.5cm,height=1.5cm]{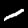}
\includegraphics[width=1.5cm,height=1.5cm]{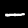}
\includegraphics[width=1.5cm,height=1.5cm]{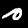}
\includegraphics[width=1.5cm,height=1.5cm]{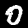}

\includegraphics[width=1.5cm,height=1.5cm]{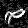}
\includegraphics[width=1.5cm,height=1.5cm]{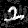}
\includegraphics[width=1.5cm,height=1.5cm]{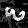}
\includegraphics[width=1.5cm,height=1.5cm]{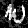}
\includegraphics[width=1.5cm,height=1.5cm]{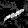}
\includegraphics[width=1.5cm,height=1.5cm]{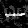}
\includegraphics[width=1.5cm,height=1.5cm]{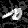}
\includegraphics[width=1.5cm,height=1.5cm]{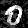}

\includegraphics[width=1.5cm,height=1.5cm]{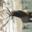}
\includegraphics[width=1.5cm,height=1.5cm]{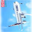}
\includegraphics[width=1.5cm,height=1.5cm]{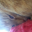}
\includegraphics[width=1.5cm,height=1.5cm]{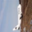}
\includegraphics[width=1.5cm,height=1.5cm]{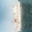}
\includegraphics[width=1.5cm,height=1.5cm]{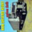}
\includegraphics[width=1.5cm,height=1.5cm]{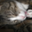}
\includegraphics[width=1.5cm,height=1.5cm]{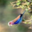}

\includegraphics[width=1.5cm,height=1.5cm]{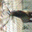}
\includegraphics[width=1.5cm,height=1.5cm]{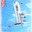}
\includegraphics[width=1.5cm,height=1.5cm]{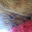}
\includegraphics[width=1.5cm,height=1.5cm]{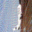}
\includegraphics[width=1.5cm,height=1.5cm]{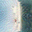}
\includegraphics[width=1.5cm,height=1.5cm]{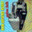}
\includegraphics[width=1.5cm,height=1.5cm]{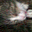}
\includegraphics[width=1.5cm,height=1.5cm]{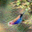}
\caption{Set of legitimate and adversarial samples from MNIST and CIFAR10 datasets: 
For each dataset, legitimate samples, which are correctly classified by DNNs, are found on the top row. 
Adversarial samples, which are misclassified by DNNs, are on the bottom row.}
\label{fig:adversarial_samples}
\end{figure*}

Specifically, the phenomenon of adversarial examples has inspired researchers to defense adversarial examples as a defender; meanwhile others play the part in attackers to design threatening attacks with slight perturbations. For example, a natural image $x$ is correctly classified by a machine learning model, while an adversarial example $x'$ is similar to $x$ but has been predicted as belonging to a different category. To produce such an $x'$, an algorithm can be designed by minimizing the perturbation and making adversarial samples imperceptible to a human (ref. Fig. \ref{fig:adversarial_samples}).

Adversarial examples are constructed to mislead the predicted output. To tackle the problem, a natural idea is to output multiple results instead of one. In this paper, we follow this idea and introduce the \textbf{Aux Block} model. In our model, a new block expands the original model; and we propose several Aux Blocks, which is like a self-ensemble model (ref. Fig. \ref{figure:aux_block}). We propose applicable Aux Block structures and show that Aux Block improves the network robustness against white-box attacks. Additionally, we cogitate the full white-box attacks (called as {\em adaptive white-box attack}), with which an adversary knows defense details thoroughly. Our algorithm can still work against the irresistible adversarial examples. 

Specifically in our work, we have made the following contributions:
\begin{enumerate}
\item We propose a novel information-appending defense method for improving the robustness of neural networks to perturbation. The key idea is to affix Aux Blocks in some convolution layers. This algorithm can be regarded as building a self-ensemble model.
\item We propose two Aux Block algorithms: a basic one and a score-based one. We show empirically that Aux Blocks could improve the robustness of model against static white-box attacks compared to the adversarial training model. 
\item We investigate the method's appearances in adaptive white-box attacks. Our method still has a robust performance even when facing with the \textbf{Adp-FGSM} attack to bypass defenses. 
\end{enumerate}

The motivation of our method is straightforward and intuitive. In practice, the method can be easily implemented and applied in different scenarios. To evaluate the performance of the proposed models, we have carried out a variety of experiments. Simple as it is, our approach has reported significantly improved results over a number state-of-the-art solutions, and therefore exhibited its high potential as a practical and effective solution to real problems.

The paper is organized as follows. Section 2 introduces the background and related work. Section 3 presents our model. Section 4 reports our experimental results, followed by conclusions and discussions in Section 5.

\section{Background}

\subsection{Neural Networks}

For a clean image $x$, denote by $f\left(\cdot \right)$ a neural network, and $y$ the class label. The feature vector $f^{i-1}$ is an input of the $i$-th layer for a feed-forward network. Activation functions such as ReLU \cite{nair2010rectified} mix in some layers to make model non-linear. Thus the $i$-th layer computes:
$$f^{i}(x)=\mathrm{ReLU}(w^{i}f^{i-1}(x)+b^i),$$ 
where $w^{i}$ is a matrix and $b^i$ is a vector. In the final layer, the predicted probability of $y_i$ is $p(y_{i}|x)$ and the predicted class is given by
$$y = \arg \max_{i} p(y_{i}|x).$$ 
The loss function of a classifier for input $x$ and target $y$ is $J\left(x,y\right)$.

\subsection{Adversarial Examples}

Adversarial examples \cite{ szegedy2013intriguing} are inputs crafted by attackers to mislead machine learning models. The malicious input $x'=x+\sigma$ and $\sigma$ is a slightly perturbation with $\|\sigma\|<\epsilon$, where $\epsilon$ is so small that it makes no visual difference between $x$ and $x'$ for human being but deep neural networks will be fooled such that $f\left(x'\right)\neq y_{true}$.

Most attacks use $L_p$-norm distance matrix to define the magnitude of $\sigma$, i.e.: $L_0$, the number of permuted pixels \cite{papernot2016limitations}; $L_2$, the Euclidean distance between $x$ and $x'$ \cite{carlini2017towards, szegedy2013intriguing, moosavi2016deepfool}; $L_\infty$, a measure of the maximum absolute change in any pixel \cite{kurakin2016adversarial}. A permuted image will be similar to clean image visually if any of three distance metrics is small.

\begin{figure*}[htb]
\centering
\includegraphics[width=15cm]{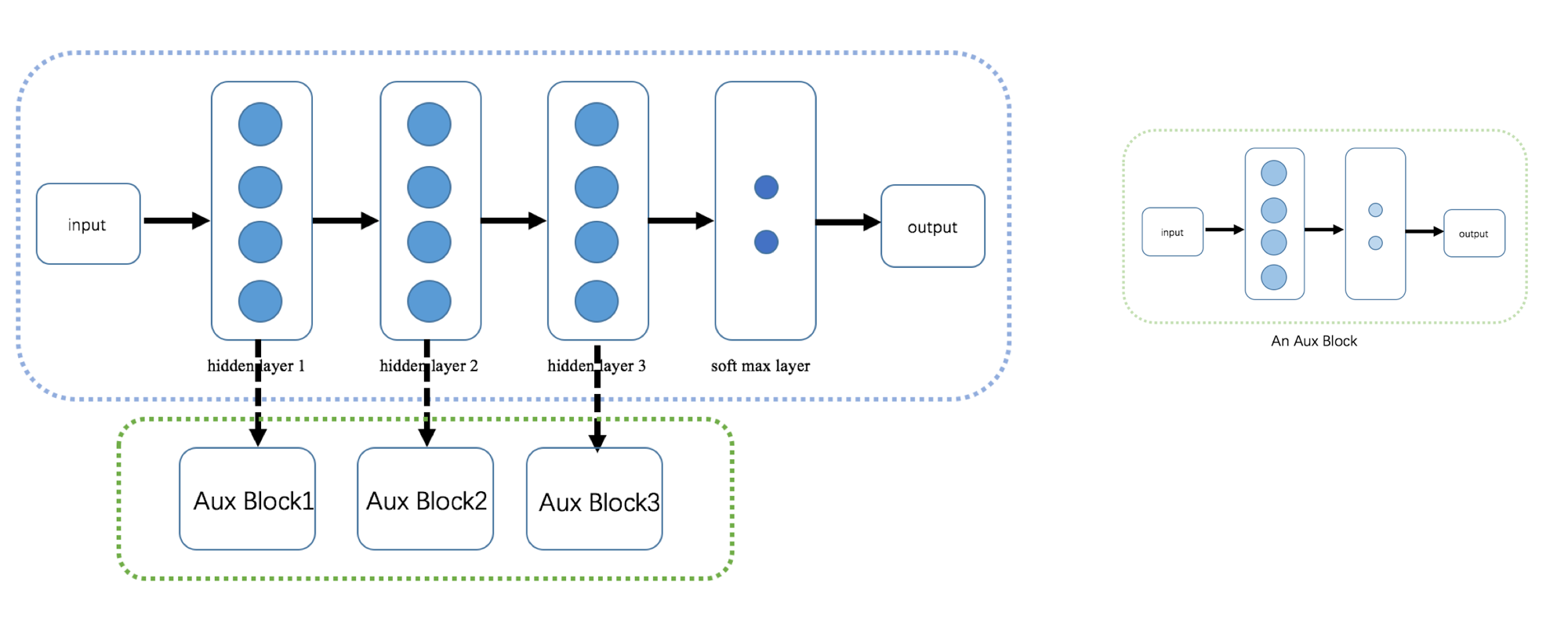}
\caption{(Left) Blue dotted circular frame is the private of an original model, while green dotted frame is the private of Aux Blocks, an Aux Block can be any function.(right) An example of an Aux Block, an one hidden layer CNN.}
\label{figure:aux_block}
\end{figure*}

Szegedy \cite{szegedy2013intriguing} first proposed to find adversarial examples in DNN models, he used L-BFGS algorithm to find them and observed that they catastrophically destroyed DNN models. They also observed that the transferability property of adversarial perturbations, one generated from an arbitrary model can also fool other models to produce incorrect outputs even if they are trained on different datasets with completely different structures. This phenomenon makes black-box attacks feasible.

Next, we discuss several attacking methods in our experiments.

\noindent \textbf{Fast Gradient Sign Method: FGSM}

Goodfellow er al. \cite{43405} proposed the FGSM method for crafting adversarial examples. FGSM is un-targeted and uses the same attack strength at every dimension:
$$x_{FGSM} = x+\epsilon \sign(\nabla_{x} J(x,y))$$
The above equation increases the loss function $J(\cdot)$ by adding a transformed gradient to input $x$, where $\epsilon$ is small enough to be undetectable.

There are several variants such as R+FGSM \cite{tramer2017ensemble}, BIM\cite{kurakin2016adversarial}, and so on.

\noindent \textbf{Projected Gradient Descent:PGD}

Madry et al.\cite{madry2017towards} reviewed adversarial robustness in a max-min saddle point problem and suggested that BIM (Basic Iterative Method applies FGSM multiple times with small step size)\cite{kurakin2016adversarial} is a projected gradient descent method essentially. In the following sections, we will replace BIM by PGD. They conjectured that Projected Gradient Descent (PGD) is the strongest attack utilizing the local first order information about the network \cite{madry2017towards}.
$$x_{N+1}=\Pi_{x+S}(x_N+\alpha(\nabla_{x} J(x,y)))$$
where $\alpha$ controls the magnitude of the perturbation in each iteration. 

\noindent \textbf{Carlini and Wagner's Attacks}

Carlini and Wagner proposed a high success rate method -- optimization-based attack algorithms \cite{carlini2017towards} to craft adversarial with low distortion. There are three versions based on different measures: $L_2$, $L_0$ and $L_\infty$. However, Carlini and Wagner's attack is computationally expensive compared to FGSM. In our paper, we select Carlini and Wagner's $L_2$ attack.

\noindent \textbf{Boundary Attacks}

Wieland et al.\cite{brendel2017decision} introduced a decision-based attack which dose not rely on gradient of the model but the final decision boundary of the model. A random perturbation are crafted at first, reduced iteratively in the adversarial region. We will use this algorithm as a gradient-free attack.

\subsection{Defenses Methods}

Adversarial Training is one of the most popular defense strategies, which is to train a model using both clean data and adversarial data \cite{ szegedy2013intriguing, 43405} to improve robustness. We study the adversarial training algorithm in \cite{madry2017towards} which solves a min-max problem:
$$\theta ^*=\arg \min_{\theta}\mathop{\mathbb{E}}\limits_{x,y\in\mathcal X} \left[\max_{\sigma\in[-\epsilon,\epsilon]^N} J(x+\sigma,y)\right ].$$
The authors solved the inner problem by crafting adversarial examples with projected gradient descent attacks.

Previous work has observed the phenomenon of \textit{gradient masking}\cite{athalye2018obfuscated}, which refers to a defense model with useless gradients by proposing a non-differentiable model \cite{buckman2018thermometer} or representing data with less details \cite{zantedeschi2017efficient}. Some defenses such as BRELU and MagNet are experimentally defeated with a small distortion\cite{carlini2017magnet}. Defensive Distillation\cite{papernot2016distillation} is also one of the most famous attack-agnostic techniques, which aims to make NN models more robust against all attacks but this method is not robust under Carlini and Wagner's attack\cite{carlini2017towards}. Moreover, most defenses obfuscate gradients somewhat, which indicate that the adversarial attack is still an unsolved problem and open for study. 

\subsection{Threat Model}

In our setting, we assume that an attacker have all knowledge of a model, such as the model parameters and model architectures in white-box attacks. Moreover, we introduce a clever attacker who is assumed to be aware of defenses that may be used.

As in Warren et al. \cite{he2017adversarial}, we consider two white-box attacks (ref. Table \ref{tbl:threat}):

\noindent \textbf{Static Adversary}

Attackers know everything about the model itself but do not realize any protection defense. A static Adversary is called partial white-box attacks in adversarial machine learning. 

\noindent\textbf{Adaptive Adversary}

An adversarial examples are crafted and tailored for specific defenses by attackers. This is more difficult to resist compared with static adversary. In this paper, we consider both static and adaptive adversaries.

\begin{table}[htb]
\caption{Threat model: adversary knows $\theta_{original}$ and/or $\theta_{defense}$}
\begin{tabular}{cp{1.5cm}cp{1cm}cp{1cm}}
\hline 
&$\theta_{original}$&$\theta_{defense}$\\
\hline  
Black-box attack&unknown&unknown\\
Static attack&known&unknown\\
Adaptive attack&known&known\\
\hline 
\end{tabular}
\centering
\label{tbl:threat}
\end{table}

\section{Method}

In this section, we illustrate our \textbf{Aux Block} model to improve neural network's robustness. We will first introduce our algorithm and motivations, and then discuss some observations behind our algorithm.

Ensemble of several different models can improve robustness against malicious inputs. However, the straits of conventional ensemble methods are also apparent: storing and reloading $k$ different models is a huge burden on memory consumption, and tuning parameters in each model is a tricky task. In this paper, we propose an algorithm that is able to generate several auxiliary classifiers in one model with less memory cost and can avoid additional parameter tuning process.

\begin{figure*}
\centering
\includegraphics[width=4cm,height=3cm]{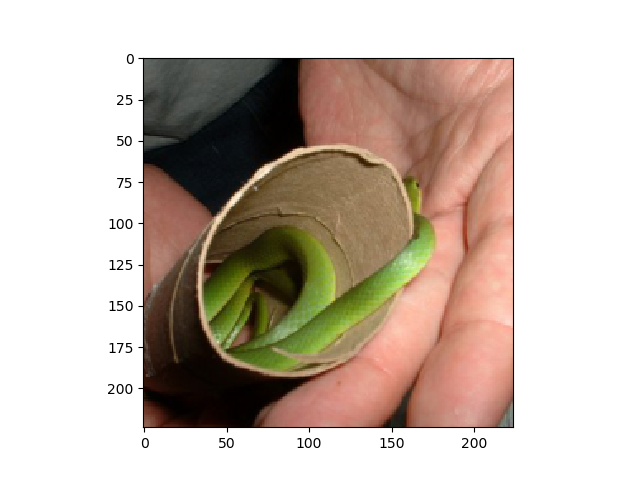}
\includegraphics[width=4cm,height=3cm]{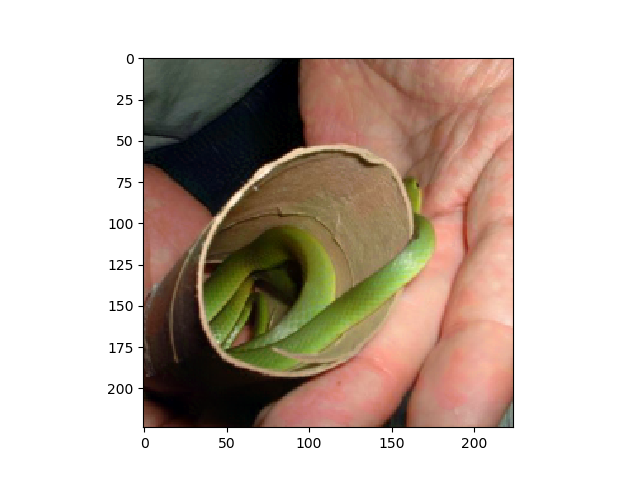}
\includegraphics[width=4cm,height=3cm]{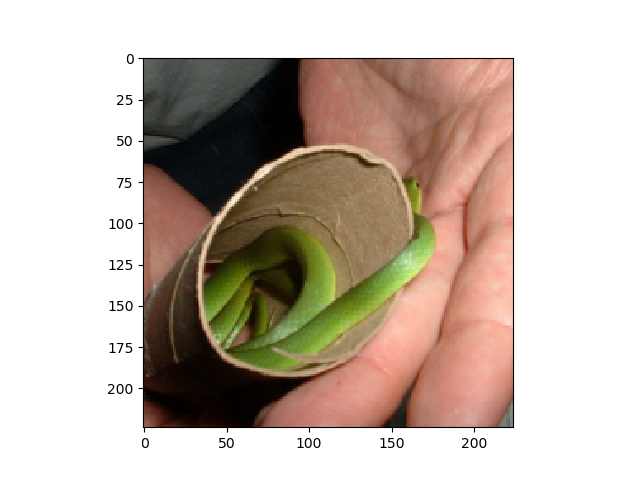}

\includegraphics[width=4cm,height=3cm]{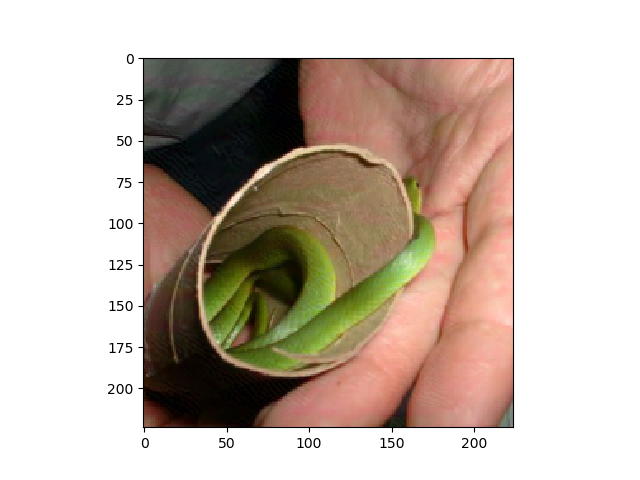}
\includegraphics[width=4cm,height=3cm]{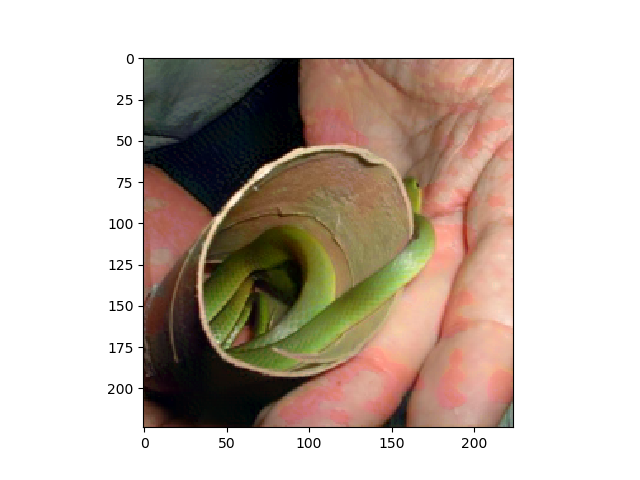}
\includegraphics[width=4cm,height=3cm]{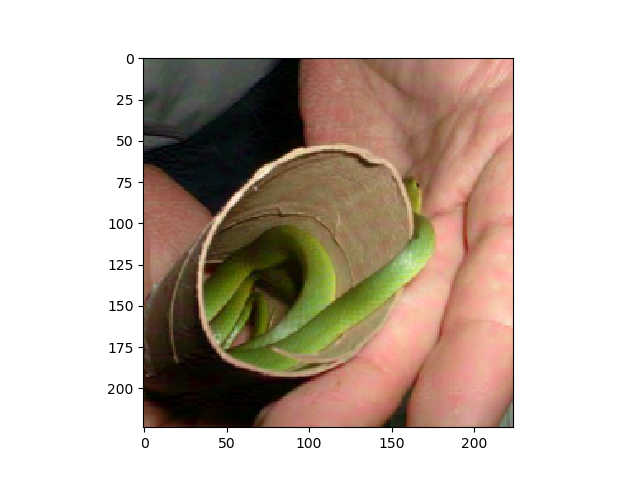}

\caption{The first row shows the adp-inputs in the non-defense model, adversarial training model and our model on Mini-Imagenet dataset. The second row shows the perturbation images from \textbf{Adp-PGD} attacks in the three models.}
\label{fig:adp_samples}
\end{figure*}

\subsection{Model}

Our main idea is to introduce \textbf{Aux Block} and then divide the model into two versions: a $public$ model and a $private$ model. More specifically, the $public$ model is a standard convolutional neural network and the $private$ is several branching auxiliary models as shown in Fig. \ref{figure:aux_block}. The $public$ model can be revealed to attackers while the $private$ model is confidential to them, so malicious hackers cannot create valid examples easily. An aux Block can be any structure, and in this paper we propose a tiny neural network.

\begin{algorithm}[htb]
\caption{Training and Testing of Aux Block}
\label{algorithm1}
\begin{algorithmic}
\REQUIRE $m$ Aux Blocks as $f_{A_1},f_{A_2},\cdots,f_{A_m}$ and an original CNN model as $f$.\\
\STATE \hspace*{-0.15in}\textbf{Training Phase:} \\
\FOR{$i=1,2,\cdots$}
\STATE Randomly sample $(x_i,y_i)$ in dataset
\STATE Compute $loss = \sum\limits_{j=1}^{m}l(f_{A_j}(x_i),y_i)+l(f(x_i),y_i)$
\STATE Compute $\Delta{w} =\nabla_{w}loss$
\STATE Update $w\gets w-\Delta{w}$\\
\ENDFOR
\STATE \hspace*{-0.15in} \textbf{Testing Phase:} 
\STATE A testing image $x$
\FOR{$j=1,2,\cdots,m+1$}
\STATE Calculate probability output of each Aux Block and the original model:
\STATE \quad $p_j = f_{A_j}(x)$, $\mathrm{if} \ j<m+1$ or $p_j = f(x)$, $\mathrm{if} \ j=m+1$
\STATE Calculate each output $y^j=\arg \max_{k}p_k $
\ENDFOR
\STATE Predict the class:
\FOR{each label $k=1,2,\cdots,n$}
\STATE $Y_k=\sum\limits_{j=1}^{m+1} \ 1$,if $y^j=k$
\ENDFOR
\STATE Predict the class $\hat{y}$,with $\hat{y}=\arg \max_k{Y_k}$. 
\end{algorithmic}
\end{algorithm}

Auxiliary network architectures are well-designed for several different datasets in the context of image classification. Details about the architectures will be discussed concretely in the next section. We denote an integral model as $f$, $x$ is the input image, and the output is an n-dimensional vector instead of a single value:
$y=[y_1,y_2,y_3,\dots ,y_{m+1}],y \ \mathrm{in} \ \Re^m $ ($m$ is the number of Aux Blocks ),  $y_1$ is the output of the $public$ network, while $y_2,y_3,\dots ,y_{m+1}$ is from Aux Blocks. Since the malevolent examples are crafted without auxiliary information, predicted outputs are less likely to be misled in Aux Blocks. Therefore our model can be deployed in security-sensitive areas.

The new cross-entropy loss function is rewritten as:
$$Loss=-\sum \limits_{i=1}^{n}\alpha_i(\sum\limits_{j=1}^{T} y_{ij} \cdot \log P_{ij}),$$
where $T$ is the number of class labels, $y_{ij}$ denotes the value of label $j$ in classifier $i$ and $P_{ij}$ is classifying possibility for $y_{ij}$. Since the new loss function is an aggregation function from each classifier, $\alpha_i \in \Re$ is a parameter which controls the relative weight of $i$ classifier in the total loss, as the confidence of an Aux Block in the final decision. In this paper, we use $\alpha=1$ for all Aux Blocks.

\begin{algorithm}[htb]
\caption{A score-based Aux Block algorithm}
\label{algorithm2}
\begin{algorithmic}
\STATE \hspace*{-0.15in}\textbf{Similar steps as Algorithm 1} 
\STATE Predict the class:
\STATE We have the Aux Block weight $\alpha_i$ and the probability of class $k$ in Aux Block $j$ as $p_{jk}$.
\FOR{$j=1,2,\cdots,m+1$ }
\STATE Calculate softmax values from probability outputs in each Aux Block
$s_j=[\frac{e^{p_{j0}}}{\sum\limits_{k=1}^{n}e^{{p}_{jk}}},\frac{e^{p_{j1}}}{\sum\limits_{k=1}^{n}e^{{p}_{jk}}},\cdots, \frac{e^{p_{jn}}}{\sum\limits_{k=1}^{n}e^{{p}_{jk}}}]$.
\ENDFOR
\STATE Sum up the probability of all Aux Blocks and the original network $s=\sum\limits_{j=1}^{m+1}{\alpha_i s_j}$.
\STATE Predict the class $\hat{y}$,with $\hat{y}=\arg \max{s}$.
\end{algorithmic}
\end{algorithm}

Our algorithm is listed in Algorithm\ref{algorithm1}. In the algorithm, we select the class $k$ which most Aux Blocks support as the final output. There are also other measures, such as a score-based algorithm with which the only difference is on selecting the final output. In Algorithm \ref{algorithm2}, we implement the same training phase but in the testing phase. We calculate values of each class and select the maximum value. Compared with Algorithm \ref{algorithm1}, Algorithm \ref{algorithm2} evaluates probability of each class more discretely and considers that an Aux Block with larger $\alpha$ has a large weight in testing. We apply Algorithm \ref{algorithm1} when $\alpha_i=1,\left( i=1,\cdots,m\right)$. Algorithm \ref{algorithm1} becomes equivalent to Algorithm \ref{algorithm2} .

\begin{figure*}[htb]
\centering
\includegraphics[width=6cm]{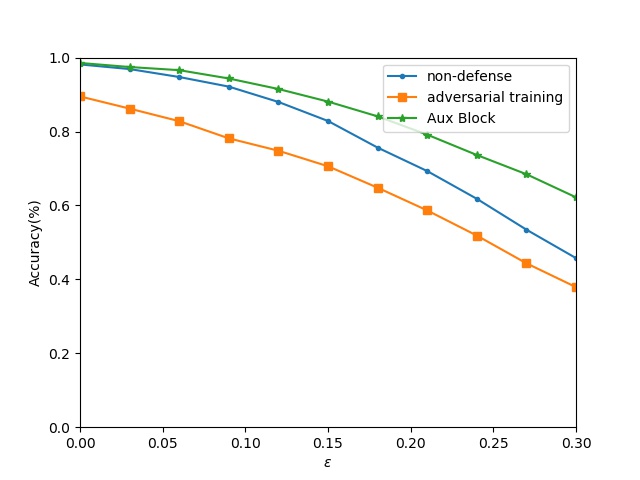}
\includegraphics[width=6cm]{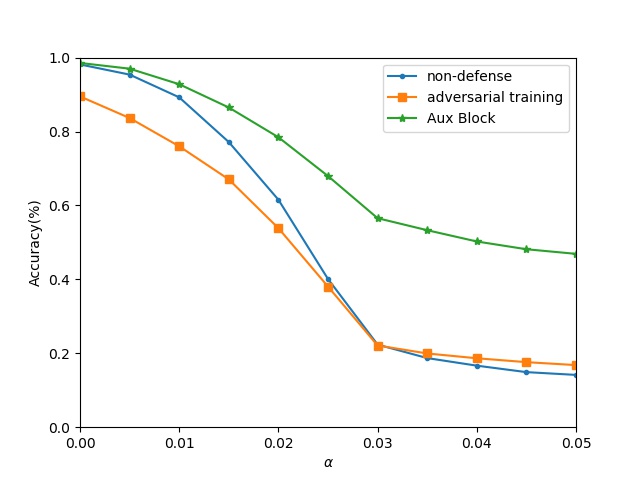}

\includegraphics[width=6cm]{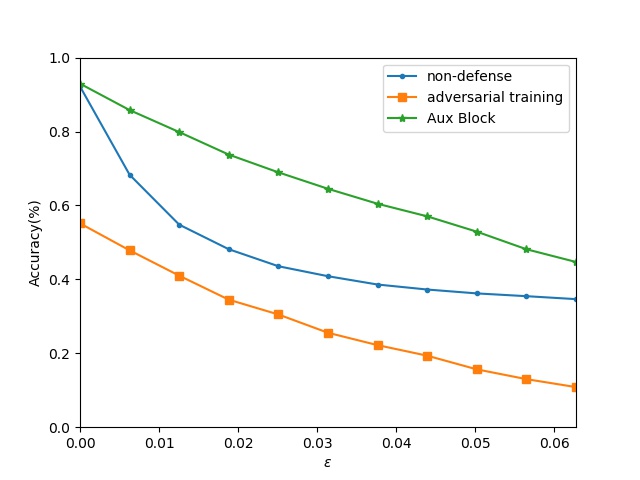}
\includegraphics[width=6cm]{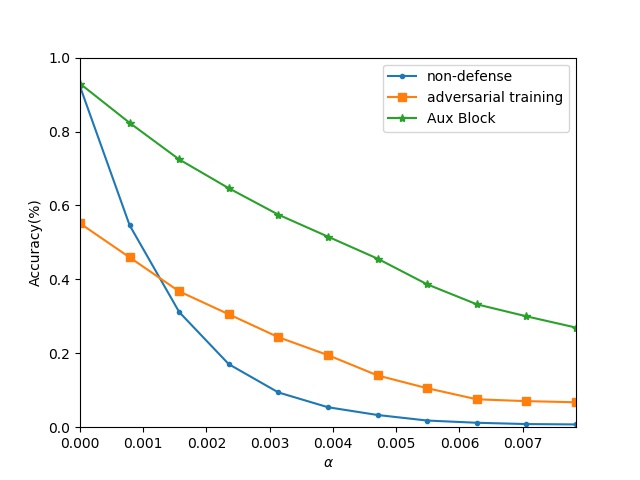}
\caption{The first row shows the accuracy curves of \textbf{Adp-FGSM} and \textbf{Adp-PGD} on MNIST dataset. The second row shows the curves on CIFAR10 dataset.}
\label{fig:figure4}
\end{figure*}

\subsection{Discussion}

In the static adversary, adversarial examples are crafted based on the non-defense model. So there can be two totally different approaches to defense these attacks. One is to reduce some necessary information, with which some existing defenses have used \cite{xu2017feature}. However, a defense performs well in a static adversary may be stock in the \textit{gradient masking} phenomenon. Anish and Nicholas \cite{athalye2018obfuscated} observed that defending models based on \textit{gradient obfuscation} is a false orientation.

Instead of information-reducing algorithms, our method increases the information by a combination output from all Aux Blocks and the original network. We do not obfuscate gradients in the original model, but an adversarial example crafted from the original model are blind to the extra knowledge in Aux Blocks. 

In the adaptive adversary, masking gradient methods are like self-deceived. When an attacker realizes the defense, he or she can bypass the defense effortless then the defense has no any effect for adversarial examples. We propose a new attack \textbf{Adp-FGSM} to craft malicious examples from defense models straight. Our method can still improve robustness in this adversary. Though an attack crafts an adversarial image for our approach, the perturbation aims to mislead one Aux Block can be right classified by other Aux Blocks. We will give more explanations about this in Section 4.4.

\section{Experiments}

\begin{table*}[htb]
\caption{Comparisons of accuracies on MNIST, CIFAR10 and Mini-Imagenet datasets,}
\begin{tabular}{cp{3cm}cp{1cm}cp{1cm}cp{1cm}c}
\hline 
Dataset&Method&Clean&PGD&CW&Boundary Attack\\
\hline
MNIST&NA&\textbf{98.88}&21.26&0.0&0.2\\
&Adversarial Training&98.76&\textbf{89.63}&58.2&\textbf{97.9}\\
&Aux Block(all)&98.49&88.2&68.4&96.6\\
&Aux Block(private)&98.21&89.59&\textbf{76.8}&97.7\\
\hline 

CIFAR10&NA&\textbf{92.53}&12.30&0.0&5.4\\
&Adversarial Training&83.89&\textbf{78.82}&76.7&83.5\\
&Aux Block(all)&91.72&74.44&77.1&87.1\\
&Aux Block(private)&89.74&77.72&\textbf{77.5}&\textbf{90.0}\\
\hline 

Mini-Imagenet&NA&\textbf{87.8}&4.6&0.0&-\\
&Adversarial Training&72.9&68.5&75.2&-\\
&Aux Block(all)&86.7&67.9&84.9&-\\
&Aux Block(private)&85.2&\textbf{71.6}&\textbf{86.0}&-\\
\hline 
\end{tabular}
\centering
\label{tbl:static}
\end{table*}

\subsection{Experiments setups}

\subsubsection{Datasets}
Our experiments are performed on three datasets: 
\begin{enumerate}
\setlength{\parsep}{0pt}
\setlength{\parskip}{0pt}
\item MNIST\cite{ lecun1998gradient} contains 70,000 samples: 60,000 training samples and 10,000 testing samples. Each sample is a grayscale image with $28\times 28$ pixels. The number of possible classes is 10.
\item CIFAR10\cite{ krizhevsky2009learning} contains 60,000 images in 10 classes: 50,000 training images and 10,000 testing images with $32\times 32$ pixels in RGB color channels.
\item We also test on Mini-Imagenet dataset with 10 classes randomly chosen from the 100 MiniImagenet classes \cite{vinyals2016matching} (each class with 600 images), on which a VGG16 model has been trained on 5000 images (500 images in each class) with 87.8 \% top-1 accuracy. 
\end{enumerate}

We consider two convolutional neural networks: LeNet5 \cite{lecun1998gradient} and VGG16 \cite{simonyan2014very}. The activation function in each network is RELU and loss function is cross-entropy function.

\subsubsection{Defense}

Adversarial training (AT) with adversarial examples crafted by PGD attacks \cite{madry2017towards}. On the MNIST dataset, we pre-train an AT model with the maximum perturbation $\epsilon=0.2$, step size $\alpha=0.02$, while on both CIFAR10 dataset and Mini-Imagenet dataset, $\epsilon$ is $8/255$ and $\alpha=2/255$.

\subsubsection{Attacking Methods}
In the static white-box adversary, we use the following attacks. 
\begin{enumerate}
\setlength{\parsep}{0pt}
\setlength{\parskip}{0pt}
\item We implement PGD with 5 iterations: $\alpha=0.1,\epsilon=0.3$ on  MNIST ,$ \alpha=2/255,\epsilon=8/255$ on CIFAR10 and $\alpha=3/255,\epsilon=10/255$ on Mini-Imagenet.
\item  We implement Carlini and Wagner's $L_2$ attack as a targeted attack.
\item We implement Boundary attacks as a gradient-free attack.
\end{enumerate}
While in the adaptive adversarial, we use the adaptive attacks. We modify FGSM and PGD to \textbf{Adp-FGSM} and \textbf{Adp-PGD} for the adaptive adversary, $x_{adv}$ are crafted from:
$$x_{adv}=x_{adp}+\epsilon(\nabla_{x_{adp}}J_{adp}(x_{adp},y)),$$
where $x_{adp}$ and $J_{adp}$ are re-designed on this particular defense.

\noindent \textbf{Adversarial Training.} We reassess that this algorithm  transforms the original inputs to adversarial inputs, therefore we use the new input $x_{adp}=x+\Delta x$, where $\Delta x$ is the perturbation for Adversarial Training.

\noindent \textbf{Aux Block.} We use the loss function defined in Section 3 to fool all Aux Blocks and the original model together: $$J_{adp}(x,y)=\sum\limits_{j=1}^{m}l(f_{A_j}(x),y)+l(f(x),y).$$

Fig.\ref{fig:adp_samples} shows the adversarial examples from \textbf{Adp-PGD}.

\begin{table}
\caption{Overview of Training Parameters: the MNIST converges after about 20 epochs; the CIFAR10 and Mini-Imagenet uses parameter decay and multi-step scheduler to ensure model convergence.}
\begin{tabular}{cccc}
\hline 
Parameter&MNIST&CIFAR10&Mini-Imagenet \\
\hline  
Learning Rate&0.01&0.01&0.01\\
Momentum&0.5&0.9&0.9\\
Decay Delay&0&0.0005&0.0005\\
Batch Size&128&128&32\\
Epochs&20&100&100\\
Multi-step Scheduler&-&0.1 (50 epochs)&0.1 (50 epochs)\\ 
\hline 
\end{tabular}
\centering
\label{tbl:hyperparameters}
\end{table}

\subsubsection{Implementation Details}

The training hyper-parameters for all models selected are in Table \ref{tbl:hyperparameters}. We use a SGD optimizer during the training phase. Especially, we run pre-trained models for 5 epochs to get adversarial training models.

\subsection{Aux Block Details}

In this section, We review VGG16 structures as [64,64,M, 128,128,M, 256 ,256, 256, M, 512, 512, 512, M, 512,512, 512, M], one Aux Block model as \textbf{Block}. 

$$\textit{The first question: Where is the suitable position of a \textbf{Block}?}\\$$

The instinct is that features in high layers are easy to be contaminated and a small perturbation in one pixel is an accumulation of the large region, since features are more abstract and the receptive field of one pixel is large. In our experiment, we insert 3 Aux Blocks in three different position: [64,64,'\textbf{Block1}', M, 128,128, M, 256,'\textbf{Block2}',256, 256, M, 512, 512, 512, M, 512,'\textbf{Block3}',512, 512, M]. We craft our Aux Block in Table \ref{tbl:s1}.

Feature maps in each convolution layer are our inputs in an Aux Block. Since the feature size shrinks after filtering each maxpooling layer, we use a flexible Aux Block structure: the number of our Aux Block layer decreases if the size of a feature map diminishes. 
\begin{table}[htb] 
\caption{\textbf{Aux Block}}
\begin{tabular}{lp{2cm}lp{2cm}}
\hline  
Layer&Parameters\\
\hline
Convolution + BatchNorm + ReLU  &$7 \times 7 \times 96$, stride=2\\
MaxPool & $2 \times 2$\\
Convolution + BatchNorm +ReLU &$3 \times 3 \times 192$\\ 
MaxPool & $2 \times 2$\\
Convolution + BatchNorm +ReLU &$3 \times 3 \times 384$\\ 
MaxPool & $2 \times 2$\\
Convolution + BatchNorm +ReLU &$3 \times 3 \times 512$\\ 
MaxPool & $2 \times 2$\\
Fully Connected &$10$\\
\hline
\end{tabular}
\centering
\label{tbl:s1}
\end{table}

As mentioned above, the three Aux Blocks have different structures: \textbf{Block1} is same as \textbf{Aux Block} in Table\ref{tbl:s1}, while \textbf{Block2,Block3} is in Table\ref{tbl:Block}. \textbf{Block3} only has a max pooling layer since the feature size is $2\times 2$.

\begin{table}[htb] 
\caption{The structure of \textbf{Block2},\textbf{Block3}.}
\begin{tabular}{lp{2cm}lp{2cm}}
\hline  
Layer&Parameters\\
\hline
MaxPool & $2 \times 2$\\
Convolution + BatchNorm +ReLU &$3 \times 3 \times 384$\\ 
MaxPool & $2 \times 2$\\
Convolution + BatchNorm +ReLU &$3 \times 3 \times 512$\\ 
MaxPool & $2 \times 2$\\
Fully Connected &$10$\\
\hline
\hline
MaxPool & $2 \times 2$\\
Fully Connected &$10$\\
\hline
\end{tabular}
\centering
\label{tbl:Block}
\end{table}

\begin{table}[htb]
\caption{The accuracy of three different \textbf{Block} in different positions, \textbf{Block1} has a highest rate.}
\begin{tabular}{ccccc}
\hline 
 \multicolumn{4}{c}{\textit{private}}&\textit{public}\\
\hline 
 &\textbf{Block1}&\textbf{Block2}&\textbf{Block3}&VGG16\\
\hline 
PGD Attack&\textbf{60.96}&53.58&53.92&53.99\\ 
\hline
\end{tabular}
\centering
\label{tbl:table2} 
\end{table}
We use a two-iteration PGD to craft adversarial examples from the $public$ model and observe each Aux Block's accuracy to investigate whether the position of an Aux Block will influence its robustness. The result in Table \ref{tbl:table2} proposes that \textbf{Block2} and \textbf{Block3} have a same performance compared to the original result, while \textbf{Block1} improves robustness to some extent. Therefore, we propose that deep layers are not a wise election for an Aux Block.

$$\textit{The second question: What is the number of Aux Blocks?}\\$$

If we create a large number of Aux Blocks, it is inescapable to select feature maps in deep layers as inputs, which will not increase the robustness of our model but aggravate the weight of incorrect output for a score-based Aux Block model and make the model become too complicated. In our experiments, we choose the first three layers to add Aux Blocks.

$$\textit{The third question: What structure is suitable for an \textbf{Aux Block}?}\\$$

A rough Aux Block will decrease the model accuracy of clean data, while the model will look overstaffed if an Aux Block is too meticulous. In this paper, we design three Aux Blocks as \textbf{Aux1},\textbf{Aux2},\textbf{Aux3} for three datasets.

For MNIST, we use the model in Table:\ref{tbl:mnist} and an Aux Bolck model in Table \ref{tbl:auxmnist}. For CIFAR10, the Aux Block structure is in Table \ref{tbl:s2}. For Mini-Imagenet, the structure is in Table \ref{tbl:mini-net}.

\subsection{Static Adversary Results}

We show our method improves the test accuracy of adversarial examples in the static adversary. We randomly select 1000 samples for the boundary attack in the testing set and do not use this algorithm in Mini-Imagenet due to the shortage of computation. 

In Table \ref{tbl:static}, our model outputs two values: one is the result from the whole system and the other is only from Aux Blocks as private results. Compared to adversarial training, Aux Block(private) has a better or equivalent performance, which illustrates that adversarial examples from the original non-defense model can be resisted by the Aux Block model. Particularly, our method's performance is better than adversarial training against Carlini and Wagner's attack on all datasets。

\subsection{Adaptive Adversary Results}

Compared with the static adversary, an adaptive adversary is a more challenging circumstance to defense. The attacker bypasses the original no-defense model and crafts specific examples from defense models straightforward.

\subsubsection{Results}

Table \ref{tbl:table5} shows the accuracy on the \textbf{Adp-PGD} attacks. In this table, our model outperforms the adversarial training model remarkably 56.55\% and 54.6\% on  Mnist and Cifar10 dataset, while adversarial training models only have about 20\% accuracy. 

\begin{table*}[htb] 
\caption{The accuracy in an adaptive adversary, with \textbf{Adp-PGD} attacks}
\begin{tabular}{ccccc}
\hline  
&\multicolumn{1}{c}{MNIST}&\multicolumn{1}{c}{CIFAR10}&\multicolumn{1}{c}{Mini-Imagenet}\\
\hline
NA&21.30&6.95&6.24 \\
Adversarial Training&22.10&22.05&25.5\\
Aux Block&\textbf{56.55}&\textbf{54.6}&\textbf{38.7}\\
\hline 
\end{tabular}
\centering
\label{tbl:table5}
\end{table*}

Fig. \ref{fig:figure4} also shows Aux Block model helps to improve the accuracy of adversarial examples. For small perturbation, even if the attacker know how to craft adversarial examples for our Aux Block model, it is still able to improve robustness compared to adversarial training models. However, as observed in the experiments, if the perturbation is large enough, our Aux Block model will also be collapsed. 

Although the experiment concerns with un-targeted attacks, it does not mean targeted attacks are not covered. As we know, targeted attacks are harder to attack and easier to defense. 

\begin{figure}[h]
\centering{}
\includegraphics[width=7cm]{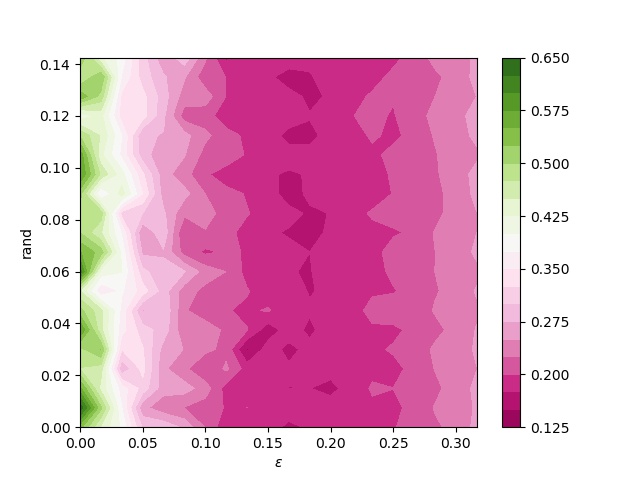}
\includegraphics[width=7cm]{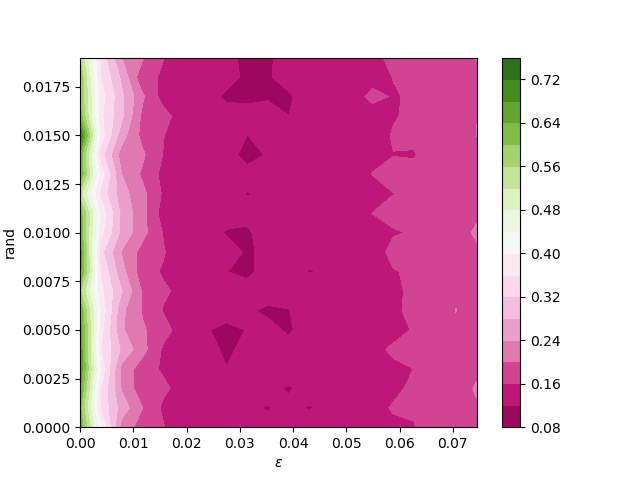}
\caption{Left: In the MNIST dataset, the ratio of the average value $l_{avg}$ to the loss value $l_i$ from the \textbf{Adp-FGSM} perturbation with a random initial. A small ratio value represents the average loss value is less than the targeted loss. Right: A figure in the CIFAR10 dataset.}
\label{fig:figure6}
\end{figure}

\subsubsection{Interpretation}

An adversarial example should fool most of Aux Blocks to mislead our method. We assume \textit{an attack for one Block is not sufficient to fool other Blocks} and \textit{crafting an example to fool all Blocks together is more difficult than fooling one}. Therefore, our model is less likely to be collapsed than non-defense one.

To prove \textit{an attack for one Block is not sufficient to fool other Blocks}, we investigate the loss value in our Aux Block model. We craft the specific perturbation targeted to one Aux Block, and then observe the loss value of this adversarial example in the selected classifier and also the average value in others. For example, the adversarial example $x_{adv}=x+\sigma_{i}$, where $\sigma_i$ is the perturbation for Aux Block $i$. Therefore, the first loss value is $l_i=l\left(f_i(x_{adv}),y\right)$ and the second value is $l_{avg}=\frac{1}{m}\sum_{k=1,k\neq i}^{m+1}l\left(f_k\left(x_{adv}\right),y\right)$.

Other auxiliary classifiers are not sensitive to the perturbation from the targeted classifier in Fig. \ref{fig:figure6}, since the ratio is less than 1 especially when the perturbation is not very large. One severe attack for an Aux Block has only slight influence to other Aux Blocks. We can see that random value do not effect the ratio. Moreover, it is obvious that \textit{crafting an example to fool all Blocks together is more difficult than fooling one}. Therefore our method is still robust in the presence of adaptive white-box adversaries.
 
\section{Conclusion and outlook}

\begin{table}[h] 
\caption{MNIST Model }
\begin{tabular}{ll}
\hline  
Layer&Parameters\\
\hline
Convolution + ReLU &$5 \times 5 \times 6$\\
MaxPool & $2 \times 2$\\
Convolution + ReLU &$5 \times 5 \times 16$\\ 
MaxPool & $2 \times 2$\\
Fully Connected &$120$\\
Fully Connected &$10$\\
\hline
\end{tabular}
\centering
\label{tbl:mnist}
\end{table}

\begin{table}[h] 
\caption{\textbf{Aux Block} in MNIST Model}
\begin{tabular}{lc}
\hline  
Layer&Parameters\\
\hline
Fully Connected &$200$\\
Fully Connected &$10$\\
\hline
\end{tabular}
\centering
\label{tbl:auxmnist}
\end{table}

In this paper, we propose a new defense technique called the Aux Block model to improve the robustness of neural networks against adversarial perturbations. We show that our algorithm is a self-ensemble model and the experimental results demonstrates that our model is robust in both two adversaries and reports significantly improved empirical results in real applications. Hence, our work lays out a new framework for adversarial machine learning. 

We suggest two lines of work for future study of the model. On one hand, more applications are expected other than the image classification applications that has been investigated in this paper. On the other hand, a theoretical analysis of the method is necessary to better understand the application scopes of the proposed model.

\begin{table}[h] 
\caption{\textbf{Aux Block} in Cifar10}
\begin{tabular}{ll}
\hline  
Layer&Parameters\\
\hline
Convolution + BatchNorm + ReLU &$9 \times 9 \times 64$ (stride=4)\\
Convolution + BatchNorm +ReLU &$3 \times 3 \times 256$\\ 
MaxPool & $2 \times 2$\\
Convolution + BatchNorm +ReLU &$3 \times 3 \times 256$ \\ 
MaxPool & $2 \times 2$\\
Fully Connected &$200$\\
Fully Connected &$10$\\
\hline
\end{tabular}
\centering
\label{tbl:s2}
\end{table}

\begin{table}[thp] 
\caption{\textbf{Aux Block} in Mini-Imagenet}
\begin{tabular}{ll}
\hline  
Layer&Parameters\\
\hline
Convolution + BatchNorm + ReLU &$9 \times 9 \times 64$ (stride=4)\\
Convolution + BatchNorm +ReLU &$3 \times 3 \times 256$ (stride=2)\\ 
MaxPool & $2 \times 2$\\
Convolution + BatchNorm +ReLU &$3 \times 3 \times 256$ \\ 
MaxPool & $2 \times 2$\\
Convolution + BatchNorm +ReLU &$3 \times 3 \times 512$ \\ 
MaxPool & $2 \times 2$\\
Fully Connected &$512$\\
Fully Connected &$10$\\
\hline
\end{tabular}
\centering
\label{tbl:mini-net}
\end{table}

\section*{Acknowledgment}

Suppressed for blind review.

{\small
\bibliographystyle{ieee}
\bibliography{egbib}

\begin{thebibliography}{10}\itemsep=-1pt

\bibitem{athalye2018obfuscated}
A.~Athalye, N.~Carlini, and D.~Wagner.
\newblock Obfuscated gradients give a false sense of security: Circumventing
  defenses to adversarial examples.
\newblock {\em arXiv preprint arXiv:1802.00420}, 2018.

\bibitem{biggio2013evasion}
B.~Biggio, I.~Corona, D.~Maiorca, B.~Nelson, N.~{\v{S}}rndi{\'c}, P.~Laskov,
  G.~Giacinto, and F.~Roli.
\newblock Evasion attacks against machine learning at test time.
\newblock In {\em Joint European conference on machine learning and knowledge
  discovery in databases}, pages 387--402. Springer, 2013.

\bibitem{brendel2017decision}
W.~Brendel, J.~Rauber, and M.~Bethge.
\newblock Decision-based adversarial attacks: Reliable attacks against
  black-box machine learning models.
\newblock {\em arXiv preprint arXiv:1712.04248}, 2017.

\bibitem{buckman2018thermometer}
J.~Buckman, A.~Roy, C.~Raffel, and I.~Goodfellow.
\newblock Thermometer encoding: One hot way to resist adversarial examples,
  2018.

\bibitem{carlini2017magnet}
N.~Carlini and D.~Wagner.
\newblock Magnet and ``efficient defenses against adversarial attacks'' are not
  robust to adversarial examples.
\newblock {\em arXiv preprint arXiv:1711.08478}, 2017.

\bibitem{carlini2017towards}
N.~Carlini and D.~Wagner.
\newblock Towards evaluating the robustness of neural networks.
\newblock In {\em 2017 IEEE Symposium on Security and Privacy (SP)}, pages
  39--57. IEEE, 2017.

\bibitem{43405}
I.~Goodfellow, J.~Shlens, and C.~Szegedy.
\newblock Explaining and harnessing adversarial examples.
\newblock In {\em International Conference on Learning Representations}, 2015.

\bibitem{he2017adversarial}
W.~He, J.~Wei, X.~Chen, N.~Carlini, and D.~Song.
\newblock Adversarial example defenses: Ensembles of weak defenses are not
  strong.
\newblock {\em arXiv preprint arXiv:1706.04701}, 2017.

\bibitem{hinton2012deep}
G.~Hinton, L.~Deng, D.~Yu, G.~E. Dahl, A.-r. Mohamed, N.~Jaitly, A.~Senior,
  V.~Vanhoucke, P.~Nguyen, T.~N. Sainath, et~al.
\newblock Deep neural networks for acoustic modeling in speech recognition: The
  shared views of four research groups.
\newblock {\em IEEE Signal processing magazine}, 29(6):82--97, 2012.

\bibitem{krizhevsky2009learning}
A.~Krizhevsky and G.~Hinton.
\newblock Learning multiple layers of features from tiny images.
\newblock Technical report, Citeseer, 2009.

\bibitem{krizhevsky2012imagenet}
A.~Krizhevsky, I.~Sutskever, and G.~E. Hinton.
\newblock Imagenet classification with deep convolutional neural networks.
\newblock In {\em Advances in neural information processing systems}, pages
  1097--1105, 2012.

\bibitem{kurakin2016adversarial}
A.~Kurakin, I.~Goodfellow, and S.~Bengio.
\newblock Adversarial examples in the physical world.
\newblock {\em arXiv preprint arXiv:1607.02533}, 2016.

\bibitem{lecun1998gradient}
Y.~LeCun, L.~Bottou, Y.~Bengio, and P.~Haffner.
\newblock Gradient-based learning applied to document recognition.
\newblock {\em Proceedings of the IEEE}, 86(11):2278--2324, 1998.

\bibitem{madry2017towards}
A.~Madry, A.~Makelov, L.~Schmidt, D.~Tsipras, and A.~Vladu.
\newblock Towards deep learning models resistant to adversarial attacks.
\newblock {\em arXiv preprint arXiv:1706.06083}, 2017.

\bibitem{moosavi2016deepfool}
S.-M. Moosavi-Dezfooli, A.~Fawzi, and P.~Frossard.
\newblock Deepfool: a simple and accurate method to fool deep neural networks.
\newblock In {\em Proceedings of the IEEE Conference on Computer Vision and
  Pattern Recognition}, pages 2574--2582, 2016.

\bibitem{nair2010rectified}
V.~Nair and G.~E. Hinton.
\newblock Rectified linear units improve restricted boltzmann machines.
\newblock In {\em Proceedings of the 27th international conference on machine
  learning (ICML-10)}, pages 807--814, 2010.

\bibitem{papernot2016limitations}
N.~Papernot, P.~McDaniel, S.~Jha, M.~Fredrikson, Z.~B. Celik, and A.~Swami.
\newblock The limitations of deep learning in adversarial settings.
\newblock In {\em Security and Privacy (EuroS\&P), 2016 IEEE European Symposium
  on}, pages 372--387. IEEE, 2016.

\bibitem{papernot2016distillation}
N.~Papernot, P.~McDaniel, X.~Wu, S.~Jha, and A.~Swami.
\newblock Distillation as a defense to adversarial perturbations against deep
  neural networks.
\newblock In {\em 2016 IEEE Symposium on Security and Privacy (SP)}, pages
  582--597. IEEE, 2016.

\bibitem{simonyan2014very}
K.~Simonyan and A.~Zisserman.
\newblock Very deep convolutional networks for large-scale image recognition.
\newblock {\em arXiv preprint arXiv:1409.1556}, 2014.

\bibitem{szegedy2013intriguing}
C.~Szegedy, W.~Zaremba, I.~Sutskever, J.~Bruna, D.~Erhan, I.~Goodfellow, and
  R.~Fergus.
\newblock Intriguing properties of neural networks.
\newblock {\em arXiv preprint arXiv:1312.6199}, 2013.

\bibitem{tramer2017ensemble}
F.~Tram{\`e}r, A.~Kurakin, N.~Papernot, I.~Goodfellow, D.~Boneh, and
  P.~McDaniel.
\newblock Ensemble adversarial training: Attacks and defenses.
\newblock {\em arXiv preprint arXiv:1705.07204}, 2017.

\bibitem{vinyals2016matching}
O.~Vinyals, C.~Blundell, T.~Lillicrap, D.~Wierstra, et~al.
\newblock Matching networks for one shot learning.
\newblock In {\em Advances in neural information processing systems}, pages
  3630--3638, 2016.

\bibitem{xu2017feature}
W.~Xu, D.~Evans, and Y.~Qi.
\newblock Feature squeezing: Detecting adversarial examples in deep neural
  networks.
\newblock {\em arXiv preprint arXiv:1704.01155}, 2017.

\bibitem{zantedeschi2017efficient}
V.~Zantedeschi, M.-I. Nicolae, and A.~Rawat.
\newblock Efficient defenses against adversarial attacks.
\newblock In {\em Proceedings of the 10th ACM Workshop on Artificial
  Intelligence and Security}, pages 39--49. ACM, 2017.

\end{thebibliography}
}

\end{document}